\documentclass{article}

\usepackage{microtype}
\usepackage{graphicx}
\usepackage{subfigure}
\usepackage{booktabs} % for professional tables

\usepackage{amsmath,amsfonts,dsfont,amssymb,color,natbib}
\usepackage{multirow}
\usepackage{hyperref}

\usepackage[accepted]{icml2021}

\DeclareMathOperator{\R}{\mathbb{R}}
\DeclareMathOperator{\N}{\mathbb{N}}

\DeclareMathOperator{\tr}{\textrm{tr}}

\graphicspath{{figs/}}

\begin{document}

\twocolumn[
\icmltitle{Spatio-Temporal Neural Network for Fitting and Forecasting COVID-19}

\begin{icmlauthorlist}
\icmlauthor{Yi-Shuai Niu}{sjtumath,sjtuparistech}
\icmlauthor{Wentao Ding}{cuhkshenzhen}
\icmlauthor{Junpeng Hu}{sjtumath}
\icmlauthor{Wenxu Xu}{sjtumath}
\icmlauthor{Stephane Canu}{insarouen}
\end{icmlauthorlist}

\icmlaffiliation{sjtumath}{School of Mathematical Sciences, Shanghai Jiao Tong University, Shanghai, China}
\icmlaffiliation{sjtuparistech}{SJTU-Paristech Elite Institute of Technology, Shanghai Jiao Tong University, Shanghai, China}
\icmlaffiliation{cuhkshenzhen}{Institute for Data and Decision Analytics, Chinese University of Hong Kong, Shenzhen, China}
\icmlaffiliation{insarouen}{Normandie University, INSA Rouen, UNIROUEN, UNIHAVRE, LITIS, France}

\icmlcorrespondingauthor{Yi-Shuai Niu}{niuyishuai@sjtu.edu.cn}

% You may provide any keywords that you
% find helpful for describing your paper; these are used to populate
% the "keywords" metadata in the PDF but will not be shown in the document
\icmlkeywords{Spatio-Temporal Neural Network; COVID-19; Stochastic Gradient Descent; Curve Fitting; Dynamical System; Supervised Learning.}

\vskip 0.3in
]

\printAffiliationsAndNotice{}  % leave blank if no need to mention equal contribution

\begin{abstract}
		We established a Spatio-Temporal Neural Network, namely STNN, to forecast the spread of the coronavirus COVID-19 outbreak worldwide in 2020. The basic structure of STNN is similar to the Recurrent Neural Network (RNN) incorporating with not only temporal data but also spatial features. Two improved STNN architectures, namely the \emph{STNN with Augmented Spatial States} (STNN-A) and the \emph{STNN with Input Gate} (STNN-I), are proposed, which ensure more predictability and flexibility. STNN and its variants can be trained using Stochastic Gradient Descent (SGD) algorithm and its improved variants (e.g., Adam, AdaGrad and RMSProp). Our STNN models are compared with several classical epidemic prediction models, including the fully-connected neural network (BPNN), and the recurrent neural network (RNN), the classical curve fitting models, as well as the SEIR dynamical system model. Numerical simulations demonstrate that STNN models outperform many others by providing more accurate fitting and prediction, and by handling both spatial and temporal data.
\end{abstract}

\section{Introduction}\label{sec:intro}
The novel coronavirus COVID-19 was firstly reported in Wuhan since December 2019. The Chinese government reacted very quickly and made great effort to limit the spread of the disease. Numerous effective control measures have been implemented, including traffic control, travel limitation, mask wearing, social distancing, party cancellation, targeted isolation, environmental disinfection, work resumption delaying, and online working etc. As a result, the epidemic situation was effectively controlled in China. By mid-February, the peak of the daily active cases appeared and dropped rapidly, and the spread of the disease in China area was almost stopped by mid-April. Meanwhile, it was just the beginning of the worldwide situation. WHO reported in April 4 that over 1 million cases of COVID-19 had been confirmed worldwide, a more than tenfold increase in less than a month. Until January 2021, there are $95$ millions cumulative cases and $2$ million deaths worldwide. Several variants of COVID-19 appeared and some of them have stronger transmission capacity, e.g., SARS-CoV-2 outbreak in UK was estimated up to 70\% more transmissible than the previously circulating forms \cite{KIRBY2021}. 

Scientists rush to understand the new illness COVID-19. Hundreds of preprints are shared every day on the study of all aspects including biomedicine, epidemiology, economics, sociology, mathematics and computer sciences etc. Several reports, e.g., \cite{world2019mers,wu2020nowcasting,leung2020nowcasting,xu2020evolution,li2020early}, show that COVID-19 has a higher basic reproductive number $R_0$ and a lower death rate comparing to the other well-known two coronaviruses SARS-CoV and MERS-CoV (SARS-CoV outbreak in 2002 caused more than $8000$ infections and $800$ deaths, and MERS-CoV outbreak in 2012 caused $2494$ individuals and $858$ deaths). Zhong's team analyzed in first the symptoms, latency and mortality of COVID-19 and emphasized the significance of early isolation \cite{guan2020clinical}. Later, the impact of different factors has been studied, such as the temperature, human mobility and control measures etc., see e.g., \cite{lin2020conceptual,chinazzi2020effect,xie2020association,kraemer2020effect}. With the spread of the pandemic to the whole world, the transmission dynamical models for different countries have been investigated, see e.g.,  \cite{phan2020importation,mizumoto2020transmission,fanelli2020analysis,rockett2020revealing,giordano2020modelling,sarkar2020modeling}.

Our work is focused on establishing a spatio-temporal deep neural network model to predict the spread of disease over time and space. There exist many works on machine learning approaches to forecast the trend of COVID-19. The study of traditional models such as linear regression, SVM for regression, fast decision tree learner and so on are reported in \cite{fong2020finding, chinazzi2020effect, ayyoubzadeh2020predicting}, which showed that the traditional methods performed well enough with small-scale dataset. The use of deep learning models such as LSTM and GRU to predict COVID-19 are reported in \cite{yang2020modified, chimmula2020time, shahid2020predictions, li2020recurrent}, which performed well in some tasks using time series dataset. However, an advanced model taking both temporal and spatial data into account to predict the spread of infectious disease is rarely appeared in literature.  

In this manuscript, we establish in Section \ref{sec:STNN} a spatio-temporal deep neural network (namely, STNN) to fit spatial and temporal pandemic data and predict the trend of the decease over time and space. Its training algorithm based on SGD is also proposed. Two improved variants of STNN, namely STNN with Augmented Spatial States (STNN-A) and STNN with Input Gate (STNN-I), are developed which could provide more flexibility and improve prediction accuracy. Some classical prediction models to compare with are briefly introduced in Section \ref{sec:othermodels} including: the fully-connected neural network model (BPNN), the recurrent neural network models (LSTM and GRU), the curve fitting (Gaussian, exponential and polynomial) models, and the SEIR dynamical system model. Our proposed STNN model and its variants are implemented in Python, and the compared classical prediction models are developed in MATLAB. These codes are tested on a set of COVID-19 dataset (a collection of various spatial and temporal data in China, United-States and Italy). Numerical results are reported in Section \ref{sec:results} which demonstrate that our STNN models perform well enough by providing accurate predictions and exhibit the ability to handle both spatial and temporal data.

\section{Spatio-Temporal Neural Network (STNN)}\label{sec:STNN}
Spatio-Temporal Neural Network, namely STNN, is a neural network architecture to predict phenomena evolving in time and in space. Thus, it is suitable for epidemic predictions.

\subsection{Classical STNN Model}\label{subsec:STNNmodel}
The classical STNN model is based on the Recurrent Neural Network (RNN) with additional spatial data. The idea to feed the neural network with spatial data comes from \cite{ziat2017spatio}. The spatial and temporal data required in STNN are described as follows:

\textbf{Model Data:} Suppose that we observe a time series data of size $m$. Temporal data at time $t\in[m]$\footnote{The notation $[m]$ with $m\in \N$ stands for $\{1, 2, \ldots, m\}$.} is denoted by $x_t \in \mathbb{R}^{n \times d}$ where $n$ is the number of observing locations (e.g., cities, countries) and $d$ is the number of observing targets (e.g., infections, deaths). The spatial data, denoted by $W_i \in \mathbb{R}^{n \times n},i\in [p]$, is a collection of $p$ matrices of spatial features (e.g., the distance between cities, the transport flow between cities). 

\textbf{Model Architecture:} The architecture of the classical STNN is illustrated in Figure \ref{fig:stnn}:
\begin{figure}[h!]
\vskip -0.1in
	\centering
	\includegraphics[width=0.8\linewidth]{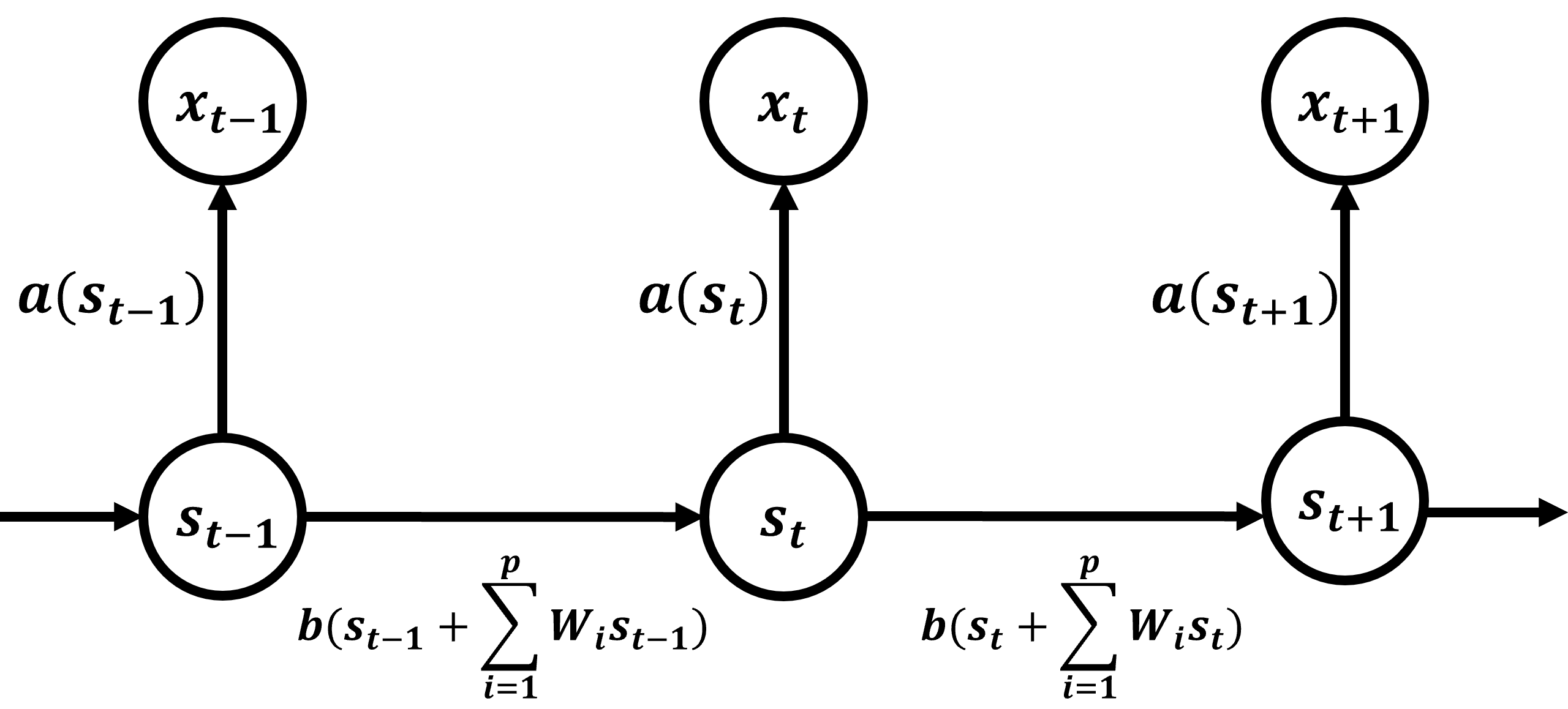}
	\caption{Structure of the classical STNN model.}
	\label{fig:stnn}
\vskip -0.1in
\end{figure}

\begin{figure}[h!]
\vskip -0.1in
	\centering
	\includegraphics[width=0.8\linewidth]{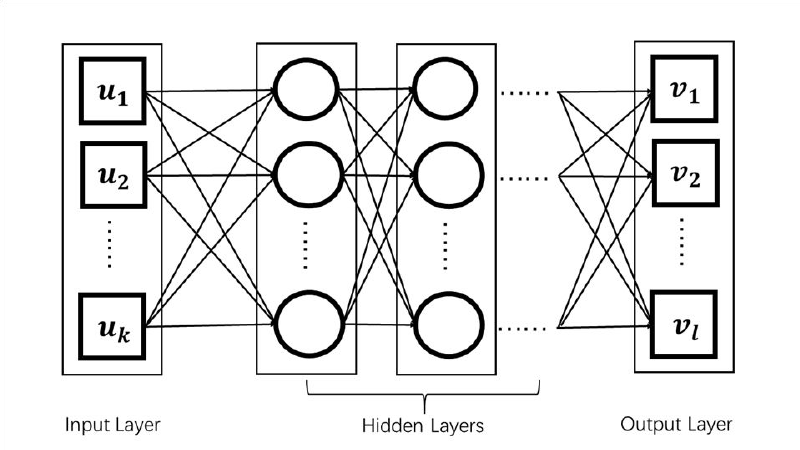}
	\caption{Common structure of the observation and state networks.}
	\label{fig:bpnn}
\vskip -0.1in
\end{figure}

STNN consists of three parts: the hidden states $s_t \in \mathbb{R}^{n \times l}, t\in [m]$ (where $l$ denotes the dimension of hidden states for each observing position), the observation network $a$ and the state network $b$. Both of the networks $a$ and $b$ are traditional fully-connected neural networks as in Figure \ref{fig:bpnn}.

The hidden state can be understood as the high-dimensional representation of the observation, and the observation is the projection of the hidden state on the low-dimensional space through the observation network. The state network is used to achieve the hidden state transition, that is, the hidden state $s_{t+1}$ is the output of the state network $b$ with input hidden state $s_t$ coupling with spatial data $W_i, \forall i\in [p]$. Due to the functionality of the observation network and the state network, we use either \verb|tanh| or \verb|sigmoid| as activation functions for all hidden layers and the output layer.

Unlike the traditional RNN model, the hidden states in STNN are also parameters to be trained, so that the loss function of STNN model is defined as:
\[
\begin{aligned}
L(\theta) =& \frac{1}{m} \sum_{t=1}^{m}\|a(s_t)-x_t\|_{F}^2 \\
&+ \frac{1}{m-1}\sum_{t=1}^{m-1}\left\| b\left( s_t + \sum_{i=1}^{p} W_i s_t\right) -s_{t+1} \right\|_{F}^2,
\end{aligned}
\]
where $\|.\|_F$ stands for the matrix Frobenius norm \footnote{Let $A$ be an $m\times n$ real matrix, the Frobenius norm of $A$ is defined by $\|A\|_F = \sqrt{\tr(A^{\top}A)}$.}, and the training parameters $\theta = (\theta_a,\theta_b,s_1,\ldots,s_m)$ in which $\theta_a$ and $\theta_b$ are parameters of the neural networks $a$ and $b$. The first summand of $L$ measures the quality of the observation network and the hidden states, and the second summand measures the quality of the state network and the hidden states. Training STNN amounts to find an observation network $a$, a state network $b$ and all hidden states $s_t, t\in [m]$ to minimize the loss function $L$, which is the following unconstrained optimization problem:
\begin{equation}
    \label{prob:STNN}
    \min_{\theta} ~ L(\theta). \tag{P}
\end{equation}

Once an ``optimal" network is obtained, the prediction is made by updating the last hidden state from $s_m$ to $s_{m+1}$ through the state network as $$s_{m+1} = b\left( s_m + \sum_{i=1}^{p} W_i s_m\right),$$ and the predicted output $x_{m+1}$ is obtained through the observation network as $a(s_{m+1})$.

\subsection{Improved STNN models}\label{sec:improvedSTNN}
There are several drawbacks of the classical STNN model. Firstly, the spatial data $W_i, i\in [p]$ are introduced by the linear mapping
\begin{equation}
    \label{eq:superposition}
    \mathcal{U}: s_t \mapsto s_t + \sum_{i=1}^{p} W_i s_t.
\end{equation}
Such a simple superposition is hard to reflect the real contribution (probably nonlinear) of each element $s_t, W_1s_t, \ldots, W_p s_t$, thus eventually limits the model's prediction accuracy and flexibility.

Secondly, there are no input data in STNN. Introducing input data $x_t$ at time $t$ may help to correct the hidden states, thus enhance the prediction accuracy.

To overcome these drawbacks, we propose two improved architectures, namely \emph{STNN with Augmented Spatial States} (STNN-A) and \emph{STNN with Input Gate} (STNN-I).

\paragraph{STNN with Augmented Spatial States}
By introducing augmented spatial states, using a separate spatial state for each spatial feature as
\begin{equation}
    \label{eq:concatenation}
    \mathcal{W}: s_t \mapsto (s_t, W_1s_t, \ldots, W_p s_t),
\end{equation}
the output of $\mathcal{W}$ is used as input of the state network $b$ which helps to provide more flexibility for improving the accuracy of the state transition.

\begin{figure}[h!]
\vskip -0.1in
	\centering
	\includegraphics[width=0.8\linewidth]{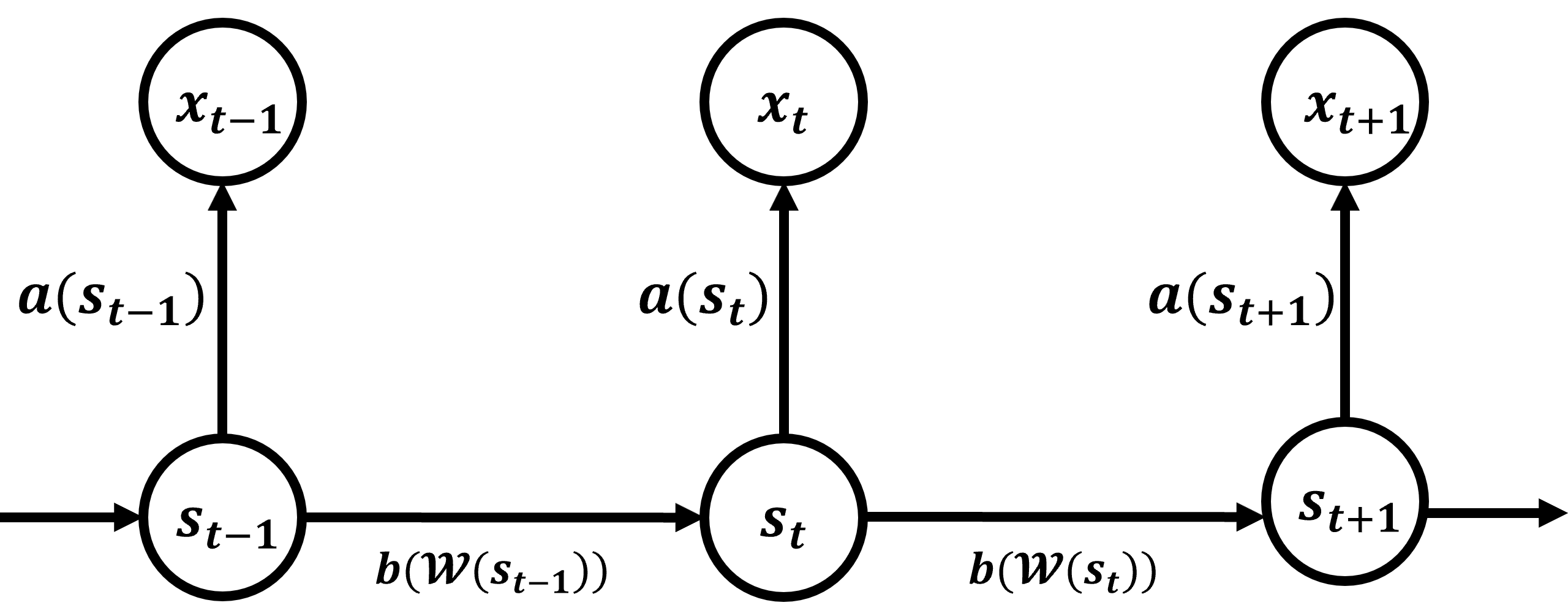}
	\caption{Structure of the STNN-A model.}
	\label{fig:stnn-augmented}
\vskip -0.1in
\end{figure}

From an algebraic point of view, both of the mappings $\mathcal{U}$ and $\mathcal{W}$ are linear. However, the mapping $\mathcal{U}$ has the same input and output dimension; while the mapping $\mathcal{W}$ has an augmented output dimension ($p+1$ times larger than the input one). This particular structure could be useful to include more rich spatial features, thereby improving the capacity of STNN to handle spatial data in a more flexible way. The improved \emph{STNN with Augmented Spatial States}, namely \emph{STNN-A model}, is described in Figure \ref{fig:stnn-augmented}.

The loss function $L_A$ for STNN-A model is defined by:
\[
\begin{aligned}
L_A(\theta) =&\frac{1}{m} \sum_{t=1}^{m}\left\|a(s_t)-x_t\right\|_{F}^2 \\
&+ \frac{1}{m-1}\sum_{t=1}^{m-1}\left\| b\left(\mathcal{W}(s_t)\right) -s_{t+1} \right\|_{F}^2.
\end{aligned}
\]
Note that both $L$ and $L_A$ have a similar structure thus can be minimized in a similar way. The prediction of STNN-A is proceed accordingly as described in STNN.

\paragraph{STNN with Inputs Gate}

In STNN and STNN-A structures, the prediction of the next time $t+1$ depends on the observation network, the state network and the current hidden state $s_t$. So if they are not accurate, the error will be propagated and accumulated through the network, and yield worse predictions over time. Therefore, we propose to introduce input data $x_{t-1}$ at time $t$ to adjust the accuracy of the hidden state $s_t$.

\begin{figure}[h!]
\vskip -0.1in
	\centering
	\includegraphics[width=0.8\linewidth]{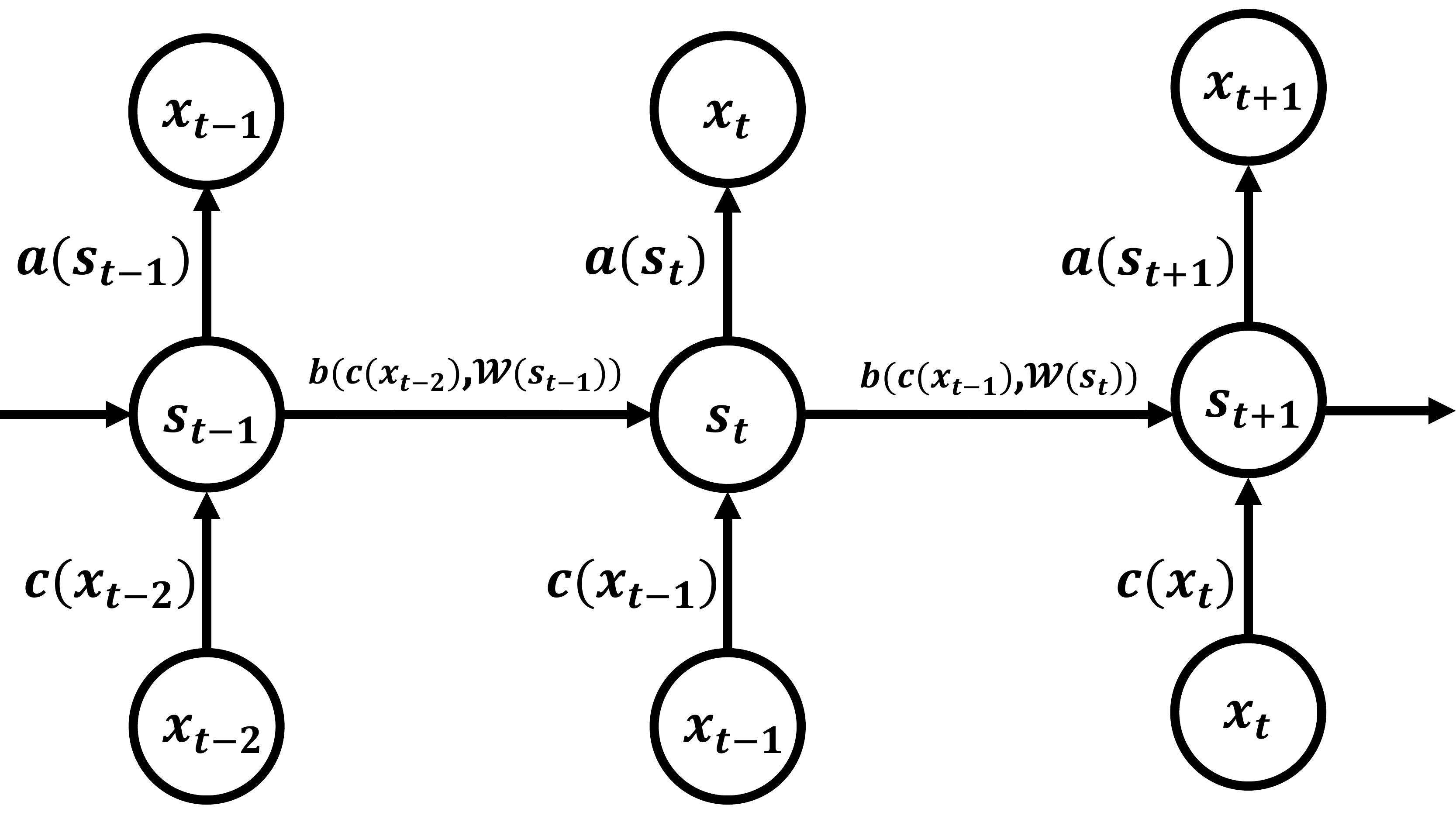}
	\caption{Structure of the STNN-I model.}
	\label{fig:stnn-input}
\vskip -0.2in
\end{figure}

To this end, the input observation $x_{t-1}$ is introduced through a fully-connected neural network, namely \emph{input network} $c$, whose output $c(x_{t-1})$ will be coupled with the hidden state $s_{t}$ at each time $t$. Then, the next hidden state $s_{t+1}$ is justified by $c(x_{t-1})$ as:
$$s_{t+1} = b(c(x_{t-1}),\mathcal{W}(s_{t})).$$
The improved STNN with Input Gate, namely STNN-I model, is described in Figure \ref{fig:stnn-input}.
Similar to STNN and STNN-A models, the loss function of STNN-I model is:
\[
\begin{aligned}
L_I(\theta) =& \frac{1}{m}\sum_{t=1}^{m}\|a(s_t)-x_t\|_{F}^2 \\
&+\frac{1}{m-2} \sum_{t=2}^{m-1}\left\| b\left(c(x_{t-1}), \mathcal{W}(s_t)\right) -s_{t+1} \right\|_{F}^2.
\end{aligned}
\]
Once the optimal network is obtained, we predict future hidden states $s_{m+t}$ as:

$$s_{m+t} = \begin{cases}
b(c(x_{m-1}),\mathcal{W}(s_{m})) &\text{, if } t=1;\\
b(c(x_{m}),\mathcal{W}(s_{m+1})) &\text{, if } t=2;\\
b\left(c(a(s_{m+t-2})),\mathcal{W}(s_{m+t-1})\right) &\text{, if } t\geq 3.
\end{cases}$$
Note that for $t\geq 3$, there is no observation data yet, thus we have to use the predicted result at time $m+t-2$, i.e., $a(s_{m+t-2})$ as input data.

\subsection{SGD and Its Variants for Training STNN Models}\label{subsec:SGD}
Training STNN models amounts to solving a large-scale nonlinear and nonconvex optimization problem of type \eqref{prob:STNN}, which is obviously a challenging problem and NP-hard in general. The most popular methods in deep learning is to use the \emph{Stochastic Gradient Descent} (SGD) algorithm and its improved variants (based on the Nesterov acceleration, Polyak momentum, adaptive learning rate, sampling techniques and noise reduction etc., see e.g. \cite{ruder2016overview,bottou1998online,bottou2008tradeoffs,Goodfellow-et-al-2016} for excellent presentations and theoretical analysis). 
Note that due to the lack of necessary and sufficient optimality conditions for nonconvex optimization, and the introduction of stochastics, SGD and its variants can be only expected to find an \emph{$\varepsilon$-stationary point}, i.e., a random vector $\theta^*$ for which $\mathbb{E}[\|\nabla L(\theta^*)\|]\leq \varepsilon$.

The term ``stochastic" in SGD is typically performed as sampling one or a minibatch of samples to compute the gradient of the loss function defined on them, namely gradient estimator $\hat{g}$ of the true gradient $g$. The gradient estimator $\hat{g}$ is supposed to be unbiased, i.e., $\mathbb{E}[\hat{g}]=g$.

Next, we will present SGD for training the classical STNN model. The variant STNN models can be trained in a similar way. Suppose that we have totally $m$ samples $\{x_1, \ldots, x_m\}$, let us choose a subset of $\hat{m}$ samples $\{x_i\}_{i\in S}$ where $S$ is the index set of samples with $S\subset [m-1]$ and $|S|=\hat{m}$. The loss function defined on $S$ is:
$$\begin{aligned}
L(\theta; S) = & \frac{1}{\hat{m}} \sum_{t\in S}\|a(s_t)-x_t\|_{F}^2 \\
&+ \frac{1}{\hat{m}} \sum_{t\in S}\left\| b\left( s_t + \sum_{i=1}^{p} W_i s_t\right) -s_{t+1} \right\|_{F}^2.
\end{aligned}$$
The classical SGD algorithm for problem \eqref{prob:STNN} is described as follows:

\begin{algorithm}[H]
	\caption{\textbf{SGD for training STNN model}}
	\label{algo:SGD}
	\begin{algorithmic}
	\STATE {\bfseries Input:} Learning rates $\{\eta_k\}$; Initial $\theta$ of STNN.
	\STATE{\bfseries Output:} Optimal $\theta$ of STNN.
	\STATE Initialize $k\leftarrow 1$;
	\WHILE{Stopping criterion not met}
        \STATE	Take a minibatch of $\hat{m}$ samples with indices in $S$;\\
        \STATE	Compute gradient estimator: $\hat{g} \leftarrow \nabla_{\theta} L(\theta;S)$;\\
        \STATE	Update parameter: $\theta \leftarrow \theta - \eta_k \hat{g}$;\\
        \STATE	$k\leftarrow k + 1;$
	\ENDWHILE
	\end{algorithmic}
\end{algorithm}

The hyper-parameter $\eta_k$ in SGD is called learning rate.  
It is necessary to gradually decrease the learning rate over time since the SGD gradient estimator introduces a source of noise (the random sampling) that does not become $0$ even at a local minimum. To guarantee the convergence of SGD, one may choose $\{\eta_k\}$ as:
$$\sum_{k=1}^{+\infty} \eta_k = +\infty~\text{and}~ \sum_{k=1}^{+\infty} \eta_k^2 < +\infty.$$
In practice, we often use the following formulation to generate a sequence $\{\eta_k\}$ as
$$\eta_k = \begin{cases} (1-\alpha_k)\eta_0 + \alpha_k \eta_\tau &\text{, if}~ 0\leq k< \tau;\\
\eta_{\tau} &\text{, if}~ k\geq \tau, \end{cases}$$
i.e., linearly decay $\eta_k$ from $\eta_0$ until iteration $\tau$ with $\alpha_k = \frac{k}{\tau}$, and fix $\eta_k=\eta_{\tau} $ when $k\geq \tau$. Concerning the choice of parameters $\tau$, $\eta_0$ and $\eta_{\tau}$, we usually set $\tau$ to the number of iterations required to make a few hundred passes through the training set; the final learning rate $\eta_{\tau}$ should be set to roughly 1\% of the initial learning rate $\eta_0$. However, the set of $\eta_0$ is a sensitive question, which should be neither too large nor too small. If it is too large, then the learning curve will show violent oscillations and cause severe instability; if it is too small, learning proceeds slowly and may become stuck with a high loss value. Typically, we set $\eta_0$ slightly greater than the learning rate that yields the best performance after the first $100$ iterations.

Concerning the convergence rate of SGD, it depends on the optimization problem we are going to solve. The convergence rate is often measured by the \emph{excess error} defined as $L(\theta^{(k)}) - \min_{\theta} L(\theta)$ where $\theta^{(k)}$ denotes the parameter $\theta$ at iteration $k$. As a result, the convergence rate of SGD is of order $O(\frac{1}{\sqrt{k}})$ for convex optimization problem, and $O(\frac{1}{k})$ for strongly convex optimization problem. These bounds cannot be improved unless extra conditions are assumed. However, training an STNN model is highly nonconvex, and the study of the convergence rate of SGD for a nonconvex optimization problem is still far from mutual. Some recent works \cite{khaled2020better,lei2019sgd,gower2019sgd} show that the optimal convergence rate for SGD to find $\varepsilon$-stationary points of nonconvex optimization problems under the expected smoothness assumption is $O(\varepsilon^{-4})$, and recover the optimal $O(\varepsilon^{-1})$ if the Polyak-Łojasiewicz condition is satisfied. Note that a training algorithm with super-linear convergence rate for machine learning is in general not expected, as \cite{bottou2008tradeoffs} argue that too fast convergence presumably corresponds to overfitting.

We have also tried some improved SGD for training STNN models, including Adam \cite{kingma2014adam}, AdaGrad \cite{duchi2011adaptive} and RMSProp \cite{tieleman2012lecture}, among which \emph{Adam} seems to outperform the others. Adam is a variant of SGD based on adaptive estimates of lower-order moments and adaptive learning rate. It includes bias corrections to the estimates of both the first and second order moments. The method is claimed to be straightforward to implement, computationally efficient, little memory requirements, well suited for problems that are large in terms of data and parameters, and appropriate for non-stationary objectives and problems with very noisy sparse gradients. The Adam algorithm for training STNN model is described in Algorithm \ref{algo:Adam} where $\circ$ denotes the Hadamard product (i.e., element-wise product). The convergence analysis of Adam can be found in \cite{reddi2019convergence}.

\begin{algorithm}[h]
	\caption{\textbf{Adam for training STNN model}}
    \label{algo:Adam}
	\begin{algorithmic}
    \STATE {\bfseries Input:} Learning rate $\eta$ ($=0.001$), exponential decay rates for moment estimates, $\rho_1$ ($=0.9$), $\rho_2$ ($=0.999$), $\delta$ ($=10^{-8}$), and initial $\theta$ of STNN.

	\STATE {\bfseries Output:} Optimal $\theta$ of STNN.
	
	\STATE Initialize $k\leftarrow 1$;
	\STATE Initialize 1st and 2nd moments $u = 0$, $v = 0$;
	\WHILE{Stopping criterion not met}
	\STATE Take a minibatch of $\hat{m}$ samples with indices in $S$;
	\STATE Compute gradient estimator: $\hat{g} \leftarrow \nabla_{\theta} L(\theta;S)$;
	\STATE Update biased 1st moment: $u \leftarrow \rho_1 u + (1 - \rho_1 ) \hat{g}$;
	\STATE Update biased 2nd moment: $v \leftarrow \rho_2 v + (1 - \rho_2) \hat{g} \circ \hat{g}$;
	\STATE Correct bias in 1st moment: $\bar{u} \leftarrow u/(1-\rho_{1}^{k})$;
	\STATE Correct bias in 2nd moment: $\bar{v} \leftarrow v/(1-\rho_{2}^{k})$;
	\STATE Update parameter: $\theta \leftarrow \theta - \eta \bar{u}/(\sqrt{\bar{v}} + \delta)$;
	\STATE $k\leftarrow k + 1$;
	\ENDWHILE
	\end{algorithmic}
\end{algorithm}

\section{Other Prediction Models}\label{sec:othermodels}
We are going to compare STNN models with several classical prediction models, including: fully-connected neural network, recurrent neural network, curve fitting models and dynamical system SEIR model. Note that these models take temporal data only. We will briefly introduce these models and explain how to use them for epidemic prediction.

\subsection{Fully-Connected Neural Network}\label{subsec:BPNN}
Fully-connected Neural Network, also called Back Propagation Neural Network (BPNN), is a popular supervised learning model, which is widely used due to its simple structure and strong fitting ability. BPNN often plays an important part of other complex neural network architectures such as CNN, ResNet, GAN as well as our STNN. 

A classical BPNN contains three parts: input layer, hidden layers and hidden layer. Every layer consists of some neurons associated with weights, bias and activation functions.  The structure of the classical BPNN is previously illustrated in Figure \ref{fig:bpnn}. 

We can use SGD and its variants to train BPNN model for epidemic prediction. The training data is a set of temporal input-output pairs $\{(x_t,y_t)\}_t$ where $x_t$ is a vector of inputs at time $t$ (e.g., the time step $t$) and $y_t$ is a vector of outputs at time $t$ (e.g., number of infections, deaths and recoveries at $t$). A well trained BPNN will receive a future input $x_t$ and output a prediction $y_t$. 

\subsection{Recurrent Neural Networks}\label{subsec:RNN}
There are two commonly used Recurrent Neural Network (RNN) architectures, namely LSTM and GRU. 

\textbf{LSTM (Long Short-Term Memory)} LSTM proposed in \cite{hochreiter1997long} is suitable for processing and predicting events with long interval and delay in time series. 
The network can choose whether to memorize or delete relevant information through a structure called "gate" (including 3 gates: forget gate, input gate, output gate). The spread of an epidemic is strongly related to the situation in a certain period, while the relation with other periods is much weaker, so that LSTM should be appropriate to predict the spread of COVID-19.

\begin{figure}[h!]
\vskip -0.1in
	\centering
	\includegraphics[width=0.8\linewidth]{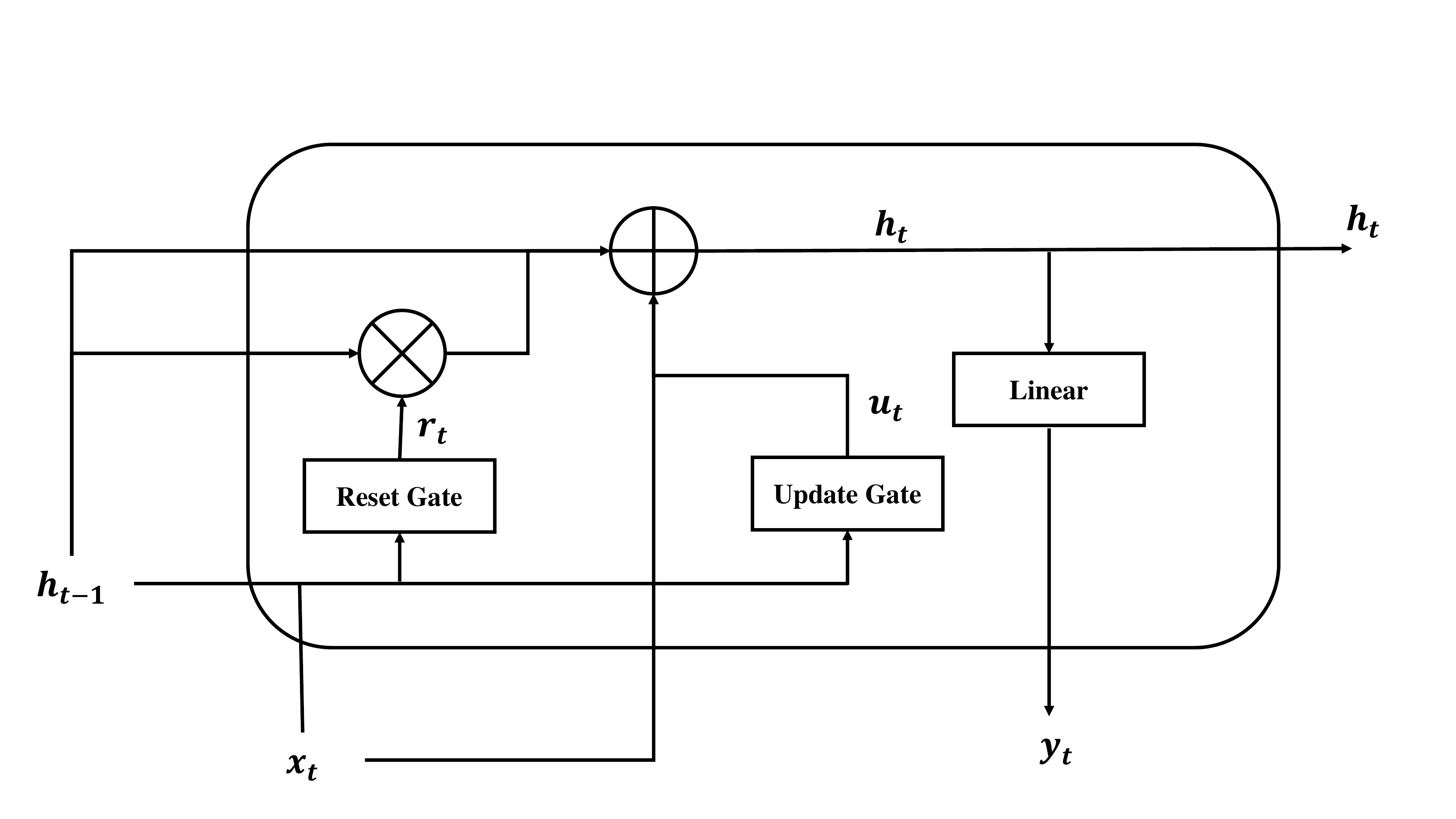}
	\caption{Structure of GRU Cell.}
	\label{fig:GRU}
\vskip -0.1in
\end{figure}

\textbf{GRU (Gated Recurrent Unit)} GRU is a variant of LSTM proposed in \cite{cho2014learning}. It is a simplified LSTM with only two gates (called update gate and reset gate) whose structure is shown in Figure \ref{fig:GRU}, thus it is claimed to be easier to train than LSTM, and can also achieve the same function as LSTM. For these reasons, GRU is now a very popular and widely used RNN. 

\begin{figure}[h!]
%\vskip -0.1in
	\centering
	\includegraphics[width=0.8\linewidth]{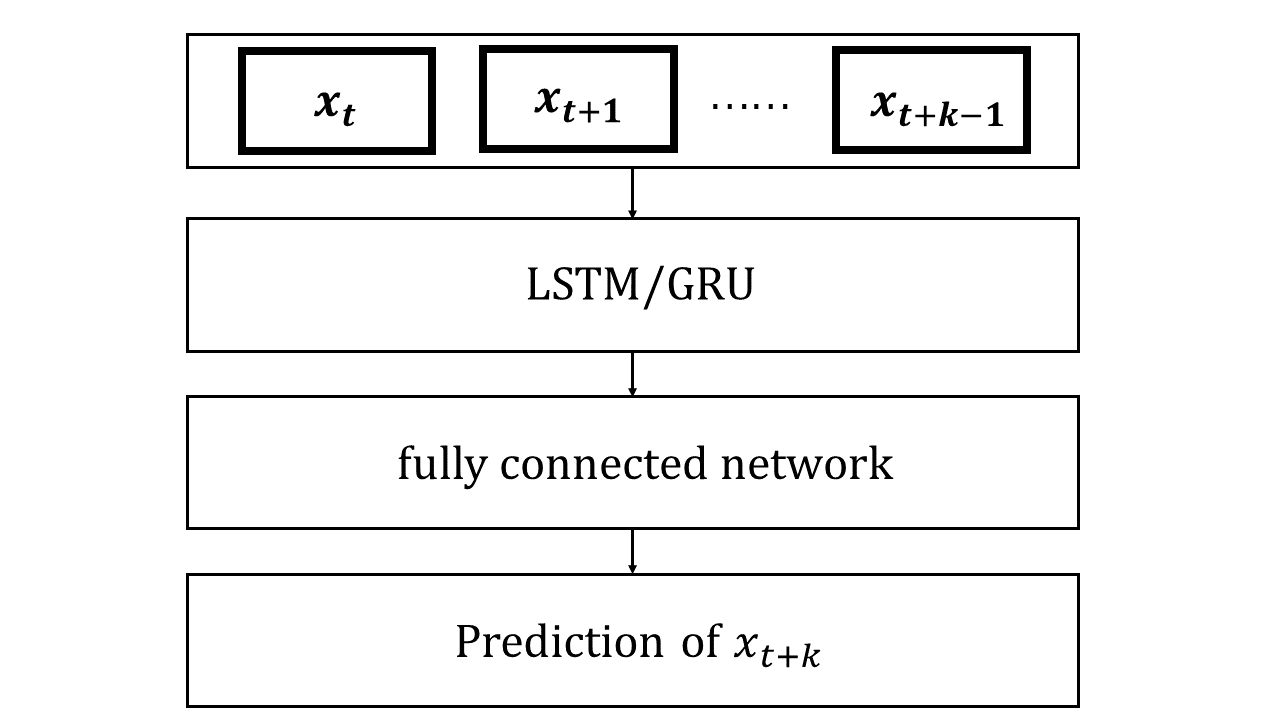}
	\caption{Prediction procedure by RNN (LSTM/GRU).}
	\label{fig:RNN}
\vskip -0.2in
\end{figure}

\textbf{Using RNN to predict COVID-19} We select a sequence of data from past $k$ time steps as input of RNN, whose output is also a sequence of the same length. Then, passing output sequence into a two layers fully-connected network (with only input and output layers) to get the prediction of the next time step. This procedure is shown in Figure \ref{fig:RNN}. Both LSTM and GRU can be trained using SGD and its variants.

\subsection{Curve Fitting Models}\label{sec:curvefittings}

Curve fitting techniques are classical prediction approaches, which aim at constructing curves, or mathematical functions, that have best fits to a series of data points. Curve fitting can involve either interpolation, where an exact fit to the data is required, or smoothing, in which a smooth function is constructed to approximate the data. Fitted curves can be used for data visualization, and to infer values of a function where no data are available, i.e., extrapolation. Therefore, we can use the fitted curves based on the observed data for epidemic prediction.

Strictly speaking, the supervised deep learning models BPNN and RNN are also curve fittings which use neural network structures (as parametric composite functions) to fit observation data. In this section, we will focus on some other type of classical curve fitting models for epidemic prediction, which can be found in many textbooks of numerical analysis (e.g., \cite{arlinghaus1994Practical}) and curve fitting packages (e.g., MATLAB Curve Fitting Toolbox).

\paragraph{Exponential Model}
The multi-terms exponential model is described by
$$y = \sum_{i=1}^{k}a_i e^{b_i x},$$
where $a_i, b_i\in \R$ are model parameters, $x\in \R$ is input, and $y\in \R$ is output.

Exponential model is often used when the rate of change of a quantity is proportional to the initial amount of the quantity. If the coefficient associated with $b_i$ is negative, $y$ represents exponential decay; otherwise, $y$ represents exponential growth. Thus, the exponential model is particularly useful to predict in the periods where the number of infections is increasing or decreasing unilaterally.

\paragraph{Gaussian Model}
The Gaussian model is often used to fit peaks defined as the sum of Gaussian functions as
$$y = \sum_{i=1}^{k} a_i e^{\left[ -\left( \frac{x-b_i}{c_i} \right)^2 \right]},$$
where $a_i$ is the amplitude, $b_i$ is the centroid, $c_i$ is related to the peak width, and $k$ is the number of peaks to fit. Gaussian fitting is particularly useful to predict peaks of the infections.

\paragraph{Polynomial Model}
Polynomial model for curve fitting is given by
$$y = \sum_{i=0}^{n}a_i x^i,$$
where $n$ is the degree of the polynomial. Polynomial model is often used when a simple empirical model is required. You can use the polynomial model for interpolation or extrapolation, or to characterize data using a global fit. In the pandemic prediction, polynomial model can be used at any period of the epidemic. The main advantages of polynomial fits include reasonable flexibility for data that is not too complicated, which means the fitting process is simple. The main disadvantage is that high-degree fits can become unstable. Additionally, polynomials of any degree can provide a good fit within the data range, but can diverge wildly outside that range. Therefore, exercise caution when extrapolating with polynomials.

\subsection{Dynamical System : SEIR Model}\label{sec:SEIR}
Differential equations are classical approaches to study epidemic dynamics involving over time. R. Ross seems to be the first to establish a mathematical differential equation to research the dynamical transmission of disease \cite{ross1911prevention}; then Kermack and Mckendrick established the warehouse SIR and SIS models \cite{kermack1927contribution,kermack1932contributions}, based on which various improved dynamical transmission models of infectious diseases have been developed such as the SIRS model, the SEIR model and the SEIRS model, see e.g., \cite{bartlett1949some,bailey1975mathematical,anderson1979population,beretta2001global}.

According to the transmission characteristics of COVID-19, patients have an incubation period, and acquire antibodies during a period after recovery. Therefore, among these dynamical transmission models, the SEIR model seems to be the most suitable one for COVID-19 prediction. The SEIR model divides persons into four categories: susceptible (S), exposed (E), infectious (I) and removed (R). The dynamics between them can be simply described as follows:
\begin{displaymath}
\begin{cases}
S'=-\dfrac{\beta SI}{N},\\
E'=\dfrac{\beta SI}{N}-\delta E,\\
I'=\delta E-\gamma I,\\
R'=\gamma I,
\end{cases}
\end{displaymath}
where $S(t)$, $E(t)$, $I(t)$, $R(t)$ are functions of the number of susceptible, exposed, infectious and removed persons over time $t$; $N$ is the total population of the area; the parameter $\beta$ is the effective contact rate, $\delta$ is the exit rate from the vulnerable to confirmed infections, and $\gamma$ is the removal rate of infected persons. These parameters can be estimated using least square method, and updated for different period to improve the fitting accuracy and get better estimations.

\section{Numerical Experiment}\label{sec:results}
In this section, we will report our numerical results for forecasting the trend of COVID-19 in several countries, including China, USA and Italy. 

Our STNN models are implemented in Python, the compared RNN (typically the GRU model is chosen instead of the LSTM model) comes from Keras, and the codes for other compared prediction models (curve fitting models, BPNN, and SEIR) are all developed in MATLAB. 
Numerical simulations are performed on a supercomputer $\pi$ 2.0, at HPC center of Shanghai Jiao Tong University, equipped with 4 GPUs (Tesla V100-SXM3) and 20 CPUs (Intel Xeon Gold 6248 CPU@2.50GHz) for neural network training.

\textbf{Data Sources}\footnote{All data are shared on GitHub at \url{https://github.com/niuyishuai/covid-19-data}.}
We collect the Chinese provincial data from National Health Commission of the People's Republic of China, the USA data from Johns Hopkins University, and Italian data from the Health Ministry. These data are all in time series, consist of the cumulative confirmed cases, deaths and recoveries.

In addition to the number of patients reported, we collect some temporal and spatial data related to the spread of the decease. The daily migration data between provinces of China are collected from Baidu Map. Some hospital information (e.g., the number of hospitals and the fever clinics) is obtained from CSMAR (short for \textit{China Stock Market $\&$ Accounting Research Database}). The population of each province is obtained from National Bureau of Statistics. Some meteorological data include temperature, humidity and air quality index (AQI) are obtained from a data service platform \textit{Nowapi}.

\begin{table}[h!]
    \caption{The correlation coefficient of different factors to the epidemic data (C: confirmed; R: recovered; D: deaths; S: sum of absolute values of CRD).}
    \label{tab:corr}
\vskip 0.15in
    \centering
	\begin{scriptsize}
	\begin{sc}
    \begin{tabular}{lcccc}
    \toprule
    
          &  C   &  R  &  D  & S\\
          \midrule
immigration   &  0.162  &  0.038  &  0.015  & 0.215\\
emigration  &  0.065  &  -0.070  &  0.002  &  0.137\\
temperature &  -0.094  &  0.033  &  -0.024  &  0.150\\
humidity      &  0.136  &  0.144  &  0.049  &  0.329\\
aqi      &  0.009  &  -0.085  &  0.026   &  0.120\\
hospitals  &  0.174  &  0.192  &  0.087 &  0.454\\
population       &  0.380  &  0.407  &  0.153 &  0.940\\
\bottomrule

    \end{tabular}
	\end{sc}
	\end{scriptsize}
\vskip -0.1in
\end{table}

A correlation analysis using provincial data (except Hubei) from January to March 2020 
is reported in Table \ref{tab:corr}. It was found that the population had the greatest impact on the epidemic situation, followed by: the number of hospitals, humidity, immigration scale index, temperature, emigration scale index and then air quality index. The main factors with major impact will be considered in STNN models.

\begin{figure}[h!]
\vskip 0.1in
\begin{center}
\centerline{\includegraphics[width=\columnwidth]{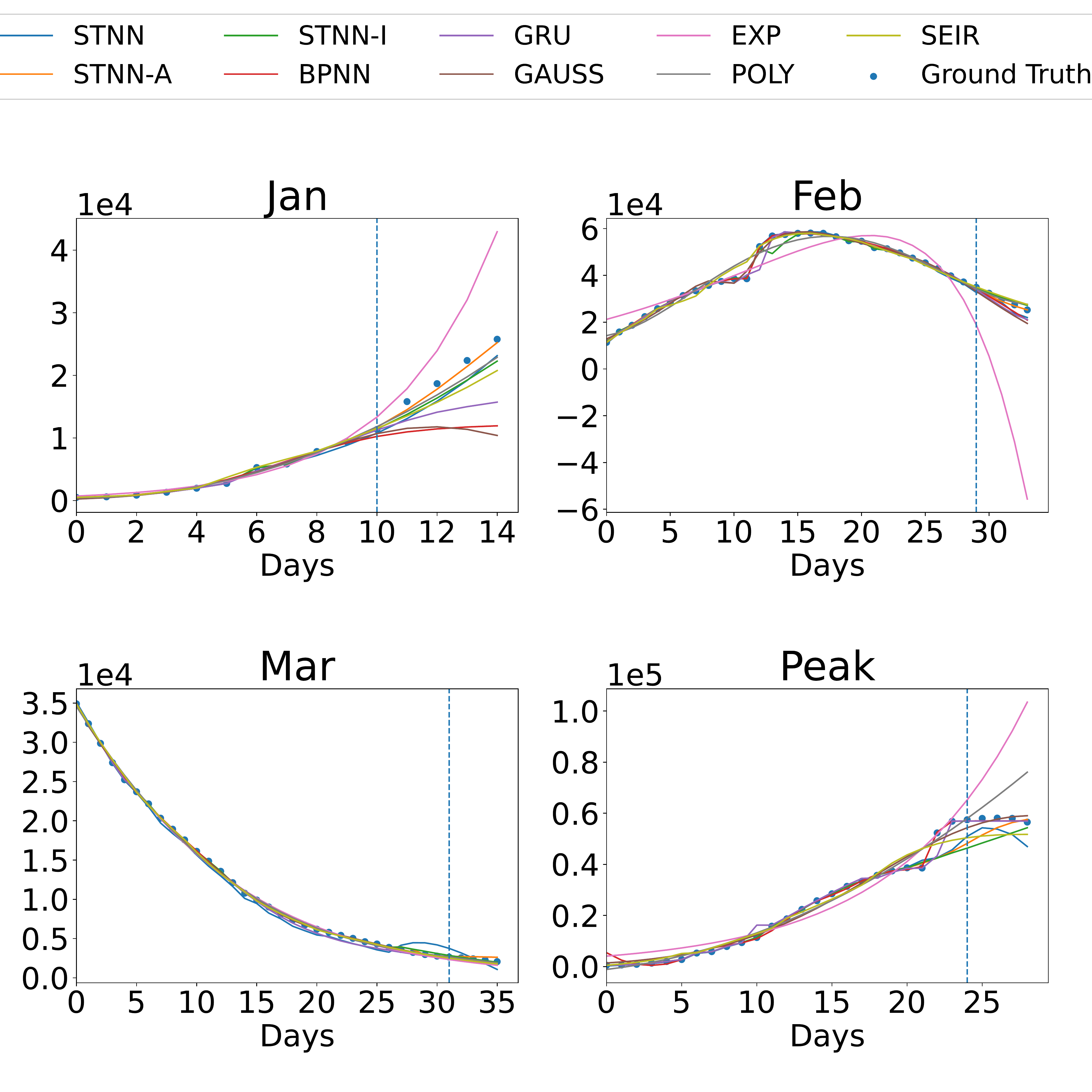}}
\caption{Fitting and prediction results of daily active cases in China using STNN, STNN-A, STNN-I, GRU, BPNN, curve fitting, and SEIR models.}\label{fig:china}
\end{center}
\vskip -0.2in
\end{figure}

\textbf{Fitting and predicting active cases for COVID-19}
Firstly, we report fitting and prediction results of daily active cases in China because it consists of a complete period from the early outbreak to the rapid control. The provincial data from January to March 2020 are used to test different models. The reason to choose this period is that the pandemic situation is very different, broke out in January, peaked in February, and reduced quickly in March. The spatial data for STNN model include the adjacency matrix of geographic locations, the average population migrations, and the distance matrix among $33$ provinces. Temporal data is divided into 2 categories: training set (90\%) and validation set (10\%). The compared BPNN, GRU, curve fitting and SEIR models are fed with temporal data only. The STNN and GRU models are trained up to 10,000 epochs using Adam algorithm, where the observation, state and input networks have $1$-$3$ layers with $10$-$30$ neurons in each. The $\ell_2$ regularization could be involved to avoid overfitting. The BPNN model is a $3$ layered network with $5$ neurons in the hidden layer and $\tanh$ as activation function, which is trained up to $1000$ epochs $10$ times using SGD with random initialization and the one with best performance is used. We apply the \emph{root mean square error} (namely, RMSE) to measure the training and prediction errors for all tested models. The fitting and prediction results using COVID-19 data in China are summarized in Table \ref{tab:rmse_china} and illustrated in Figure \ref{fig:china}.
\begin{table}[h!]
\caption{Training and testing RMSE (base on $10^3$) of daily active cases in China, boldface/underline for best/worst records.}
\label{tab:rmse_china}
\vskip 0.15in
    \centering
    \begin{scriptsize}
	\begin{sc}
	\resizebox{\columnwidth}{!}{
    \begin{tabular}{lcccccccc}
\toprule
          &  \multicolumn{2}{c}{Jan.}   &  \multicolumn{2}{c}{Feb.}  &  \multicolumn{2}{c}{Mar.} &
          \multicolumn{2}{c}{Peak.}\\
          & Train & Test & Train  & Test &  Train &  Test & Train & Test  \\
\midrule
STNN   &       0.27 &      2.56 &       0.46 &      2.47 &       \underline{0.68} &      \underline{0.75} &        3.09 &       6.43 \\
STNN-A &       \textbf{0.00} &      \textbf{0.87} &       \textbf{0.23} &      \textbf{0.83} &       0.09 &      0.34 &        3.15 &      5.40 \\
% STNN-A &       \textbf{0.18} &      \textbf{1.76} &       \textbf{0.23} &      \textbf{0.83} &       0.09 &      0.34 &        3.15 &      5.40 \\
STNN-I &       0.02 &      2.51 &       1.56 &      1.10 &       0.13 &      \textbf{0.11} &        3.26 &       7.93 \\
BPNN   &       0.32 &      8.73 &       0.36 &      2.65 &       \textbf{0.08} &      0.36 &        \textbf{1.24} &       \textbf{0.85} \\
GRU    &       0.22 &      6.08 &       1.95 &      3.19 &       0.36 &      0.29 &        2.12 &       0.90 \\
GAUSS  &       0.31 &      9.20 &       1.16 &      4.06 &       0.14 &      0.37 &        2.62 &       1.98 \\
EXP    &       \underline{0.51} &      \underline{9.21} &       \underline{5.18} &     \underline{50.25} &       0.27 &      0.45 &        3.79 &      \underline{29.13} \\
POLY   &       0.32 &      2.06 &       2.65 &      1.23 &       0.23 &      0.12 &        2.32 &      11.42 \\
SEIR   &       0.41 &      3.39 &       1.86 &      1.49 &       0.15 &      0.27 &        2.91 &       6.38 \\
\bottomrule
    \end{tabular}}
	\end{sc}
	\end{scriptsize}
\vskip -0.1in
\end{table}

\begin{figure}[h!]
\vskip 0.1in
\begin{center}
\centerline{\includegraphics[width=\columnwidth]{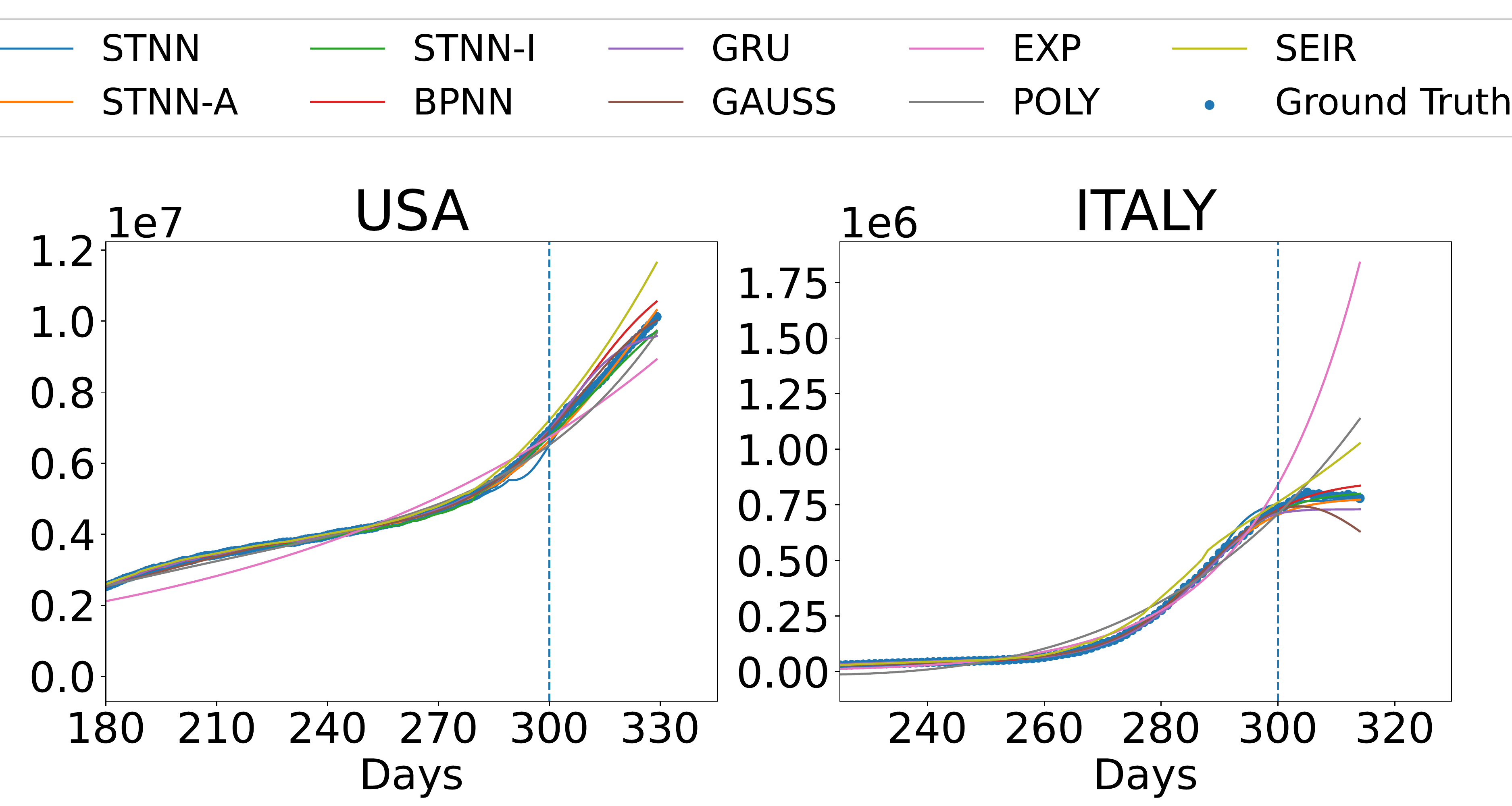}}
    \caption{Fitting and prediction results of daily active cases in USA and Italy using STNN, STNN-A, STNN-I, GRU, BPNN, curve fitting, and SEIR models.}\label{fig:global}
\end{center}
\vskip -0.2in
\end{figure}

Concerning global outbreaks outside of China, we choose two typical countries: USA and Italy. The fitting and prediction results are reported in Figure \ref{fig:global}; the training and testing errors are summarized in Table \ref{tab:rmse_global}. The first 300 days of spatial and temporal data from 22 January to 31 December of 2020 are used to predict the peak of the next 15 days in Italy and the trend of continued growth over the next 30 days in USA.

\begin{table}[h!]
\caption{Training and testing RMSE (base on $10^4$) of daily active cases in USA and Italy, boldface/underline for best/worst records.}
\label{tab:rmse_global}
\vskip 0.15in
    \centering
    \begin{scriptsize}
	\begin{sc}
   \begin{tabular}{lcccc}
\toprule
          &  \multicolumn{2}{c}{USA}   &   \multicolumn{2}{c}{Italy} \\
          & Train & Test & Train  & Test  \\
\midrule
STNN   &      14.00 &     20.79 &         0.83 &        2.26 \\
STNN-A &       4.84 &     19.93 &         0.33 &        3.52 \\
STNN-I &       6.75 &     23.59 &         \textbf{0.19} &        \textbf{1.92} \\
BPNN   &       1.48 &     42.36 &    \textbf{0.19} &        2.59 \\
GRU    &       \textbf{1.05} &    27.08 &      0.60 &        5.87 \\
GAUSS  &      12.22 &     \textbf{16.35} &    0.53 &        8.44 \\
EXP    &      \underline{38.08} &     76.27 &   \underline{4.49} &       \underline{57.99} \\
POLY   &      15.14 &     56.30 &    3.08 &       17.79 \\
SEIR   &       7.83 &     \underline{86.92} &        1.58 &       12.92 \\

\bottomrule
    \end{tabular}%}
	\end{sc}
	\end{scriptsize}
\vskip -0.1in
\end{table}

\textbf{Comments on numerical results} Numerical results demonstrate that the STNN-A, STNN-I and BPNN models often provide best fitting and prediction results (as shown in Figures \ref{fig:china} and \ref{fig:global} and in Tables \ref{tab:rmse_china} and \ref{tab:rmse_global} with boldface for smallest RMSE and underline for largest RMSE).  
The classical STNN, GRU, SEIR and polynomial models often perform well for both fitting and prediction of data in China and other countries, only slightly weaker in RMSE than the improved STNN models and BPNN model. The exponential model seems to perform bad for both fitting and prediction, even for unilateral cases, and not suitable to fit peaks; while BPNN, GRU, STNN and Gaussian models perform quite well to predict peaks. The major drawback of the SEIR model is the high dependence on the estimation of model parameters; the Gaussian model performs well for unilateral fitting and short period unilateral prediction, but not very satisfactory for long period unilateral prediction; the polynomial model works well for fitting within the data range, but diverges wildly for prediction beyond that range. As for our STNN models, the main advantage is the compatibility and the flexibility of using both temporal and spatial data to increase the accuracy (with small RMSE) of fitting and prediction; while its main drawback is also the requirement of many types (temporal and spatial) of data whose collection could be difficult and cumbersome. As plus, the training of deep neural networks (e.g., STNN and RNN) is very time and computing resource consuming comparing to the other models. Nevertheless, there is a need of large amounts of training data in many deep learning models, whose collection, in the current era of big data, is not too difficult. The increasing GPU computing power also makes it possible to train deep neural networks. Therefore, we believe that STNN should be a promising deep learning model in many potential applications involving spatial and temporal data. 

\section*{Acknowledgments}
This project is partially supported by the special grand for Science and Technology Innovation at Shanghai Jiao Tong University ``Spatio-Temporal Deep Neural Network for Trend Prediction of the New Coronavirus COVID-19 and Countermeasures Researches'' (Grant 2020RK10), 2020, and by the National Natural Science Foundation of China (Grant 11601327).

\bibliography{references}
\bibliographystyle{icml2021}
\end{document}